\documentclass[letterpaper, 10 pt, conference]{ieeeconf}  %

\IEEEoverridecommandlockouts                              %

\usepackage{amsmath,amsfonts,amssymb}
\usepackage[protrusion=true,spacing=true]{microtype}
\usepackage[bookmarks=true]{hyperref}
\usepackage[noend]{algorithmic}
\usepackage{algorithm}
\usepackage{graphicx}
\usepackage{setspace}
\usepackage[font=small]{caption}
\usepackage{subcaption}
\usepackage{tabularx}
\usepackage{multirow}
\usepackage[T1]{fontenc}
\usepackage{lscape}
\usepackage{setspace}
\usepackage[capitalise]{cleveref}
\usepackage[dvipsnames]{xcolor}

\DeclareMathOperator*{\argmax}{arg\,max}

\newcommand{\dash}{{\text -}}

\title{\LARGE \bf
Verifiable Goal Recognition for Autonomous Driving with Occlusions
}

\author{\authorblockN{
        Cillian Brewitt\authorrefmark{1},
        Massimiliano Tamborski\authorrefmark{1},
        Cheng Wang\authorrefmark{1},
        Stefano V. Albrecht\authorrefmark{1}\authorrefmark{2}} 
        \texttt{\{cillian.brewitt@, m.tamborski@sms., cheng.wang@, s.albrecht@\}ed.ac.uk}
        \authorblockA{\authorrefmark{1}School of Informatics, University of Edinburgh, UK  \hspace{1.0pt} \authorrefmark{2}Five\,AI Ltd. / Bosch, UK }%
        \vspace{-1.0em}
}

\begin{document}

\maketitle
\thispagestyle{empty}
\pagestyle{empty}

\begin{abstract}

    Goal recognition (GR) involves inferring the goals of other vehicles, such as a certain junction exit, which can enable more accurate prediction of their future behaviour. In autonomous driving, vehicles can encounter many different scenarios and the environment may be partially observable due to occlusions. We present a novel GR method named Goal Recognition with Interpretable Trees under Occlusion (OGRIT). OGRIT uses decision trees learned from vehicle trajectory data to infer the probabilities of a set of generated goals. We demonstrate that OGRIT can handle missing data due to occlusions and make inferences across multiple scenarios using the same learned decision trees, while being computationally fast, accurate, interpretable and verifiable. We also release the inDO, rounDO and OpenDDO datasets of occluded regions used to evaluate OGRIT.

\end{abstract}

\section{Introduction}

To navigate through complex urban environments, an autonomous vehicle (AV) must have the ability to predict the future trajectories of other road users. One method of doing this is to first perform goal recognition (GR) to infer the goals of other vehicles, such as taking certain junction exits or roundabout exits. If the goals of road users are known, this can facilitate the prediction of their trajectories over longer horizons \cite{rhinehart_precog_2019,mangalam_it_2020,zhao_tnt_2020}. For example, a planner can be used to plan multiple possible trajectories to an inferred goal, which can then be used for trajectory prediction \cite{albrecht_interpretable_2021}.

GR is a difficult task, as there are many criteria that GR methods must fulfil to be applied successfully to autonomous driving. GR methods must be able to handle \textbf{missing information} due to occlusions. It is also important that GR methods have the ability to \textbf{generalise} across many scenarios. Inferences made by GR methods must be \textbf{accurate} to be useful for planning, and they must be \textbf{fast} so that they can run in real time. GR methods should ideally be \textbf{interpretable}, which can improve user trust and help with the debugging of the method. As autonomous driving is a safety-critical domain, it is also important that the inference process is \textbf{verifiable}. This can be achieved by formally proving that certain statements made about the method will hold true under all possible conditions \cite{butler_infeasibility_1993,Luckcuck_formal_2019,schmidt_can_2021}. Existing GR methods for AVs do not satisfy all of the above requirements.

We present a novel GR method named \textit{Goal Recognition with Interpretable Trees under Occlusion} (OGRIT). OGRIT\footnote{{\bf OGRIT code, video, and inDO, rounDO and OpenDDO datasets:} \url{https://github.com/uoe-agents/OGRIT} \label{fn1}} uses decision trees (DTs) trained with vehicle trajectory data to infer the likelihoods of goals. To handle missing data, OGRIT uses indicator features which show when certain DT input features are missing due to areas being occluded. These indicator features are added to the DT using a novel training algorithm which uses a one-level look-ahead. One DT is trained for each goal type, such as exiting left at a junction, and the same DT can be used when a goal of that type is encountered across multiple scenarios. Each specific goal location in the scenario is assigned a goal type depending on the maneuvers the ego must execute to reach that goal from its current position.

We evaluate OGRIT across 11 scenarios from three different vehicle trajectory datasets, inD~\cite{bock_ind_2019}, rounD~\cite{krajewski_round_2020}, and OpenDD~\cite{openDD_dataset}. We show that OGRIT can handle occlusions and generalise across multiple scenarios, while being computationally fast, accurate, and interpretable. We also formally verify that certain propositions about predictions made by the models learned by OGRIT will always hold true. For example, we verify that if a certain input feature is missing due to occlusions, the inferred goal distribution entropy will be greater than or equal to the entropy if the feature is not missing, all other features being equal.

As the original datasets are not already annotated with occluded regions, we developed a tool to extract occluded regions in 2D trajectory data. We release the tool and the extracted occlusion datasets\footref{fn1}, which we name inDO, rounDO and OpenDDO. Along with the original datasets, inDO, rounDO and OpenDDO can serve as a benchmark for AV systems designed to handle occlusions.

{\bf In summary, our main contributions are:}
\begin{itemize}
  \item The inDO, rounDO and OpenDDO datasets of occluded regions, which annotate three existing vehicle trajectory datasets.
  \item OGRIT, a novel GR method that handles occlusions and is generalisable, fast, accurate, interpretable and verifiable.
  \item A comprehensive evaluation and analysis  of OGRIT using the inDO, rounDO, and OpenDDO datasets.

\end{itemize}

\section{Related Work}

Many existing AV prediction methods make use of deep neural networks~\cite{rhinehart_precog_2019,lee_desire_2017,chai_multipath_2019,Knittel2023dipa,casas_intentnet_2018,christianos2023planning,antonello2022flash}. Although such methods can achieve high accuracy, neural networks are black box models that are not easily interpretable. In addition to this, neural networks often have millions of learned parameters, making verification intractable~\cite{ayers_parot_2020}. Another approach to prediction is to first perform GR through inverse planning~\cite{hardy_contingency_2013,bandyopadhyay_intention-aware_2013,ziebart_maximum_2008,darweesh_estimating_2019}. A recent method named IGP2 \cite{albrecht_interpretable_2021} uses an interpretable inverse planning approach to GR, which can handle occlusions by using a planner to fill gaps in trajectories. Agents are assumed to be rational, and unobserved sections of trajectories are imputed with the optimal plan. A similar GR method named GOFI~\cite{hanna_interpretable_2021} uses inverse planning to infer occluded factors. Inverse planning-based methods can be accurate, interpretable and handle occlusion. However, these methods are typically slow and unverifiable due to the complexity of inverse planning.

A recent method named GRIT~\cite{brewitt_grit_2021} makes use of decision trees trained on vehicle trajectory data to perform GR for AVs. GRIT is shown to be fast, and relatively accurate compared to deep learning and inverse planning-based methods. The inference process of GRIT is also highly interpretable due to the shallow depth of the trees and interpretable input features used. The properties of the trained GRIT models can also be formally verified due to the simplicity of DT inference. However, GRIT assumes that all vehicles in the local area are fully observable when, in reality, some of the DT input features can have missing values due to occlusions. In addition to this, the GRIT models are trained specifically on the goal positions for a given fixed scenario and do not readily generalise across different scenarios.

Many existing methods allow DT inference with missing data~\cite{gavankar_decision_2015}. Some methods compute the expected inference based on the distributions over the values of missing features~\cite{khosravi_handling_2020, quinlan_induction_1986}. However, such methods introduce more complexity to the DT inference process. This reduces their interpretability and may make such methods unverifiable. Another approach is lazy decision trees~\cite{friedman_lazy_1997}, where a new DT is trained for each inference made. However, this is computationally expensive and is not suitable when inferences are needed in real-time. As part of our work, we instead introduce a novel DT training algorithm designed to handle missing feature values while resulting in trained DTs which have fast, interpretable and verifiable inference.

\section{Problem Definition}

Let $\mathcal{I}$ represent the local set of vehicles surrounding the ego vehicle under our control. The state of vehicle $i \in \mathcal{I}$ at time $t$ is represented by $s^i_t \in \mathcal{S}^i$, and includes information about the pose, velocity and acceleration of the vehicle. The state of all vehicles at time $t$ is given by $s_t = \{s_t^i | i \in \mathcal{I} \} \in \mathcal{S}$. At each time $t$, the ego vehicle receives an observation $o_t \subseteq s_t$ which contains the state of a subset of vehicles. The state of some vehicles may be missing from $o_t$ due to occlusions. The observations between times $a$ and $b$ are denoted by $o_{a:b}=\{o_a,o_{a+1},...,o_b\}$. We assume a set of $k$ possible goals $G^i_t = \{g^{i,1}_t, g^{i,2}_t, ..., g^{i,k}_t\}$ that each vehicle $i$ can reach at time $t$, where a goal is a subset of states $g_t^{i,k} \subset \mathcal{S}$ such as a target location in a given scenario. We define the problem of goal recognition under occlusion as the task of inferring the probability distribution over goals based on past observations, $P(g_t^{i,k}|o_{1:t},\phi)$, where $\phi$ represents static scene information such as the road layout and static obstacles.

\begin{figure}[t] 
  \centering
  \includegraphics[width=0.75\linewidth]{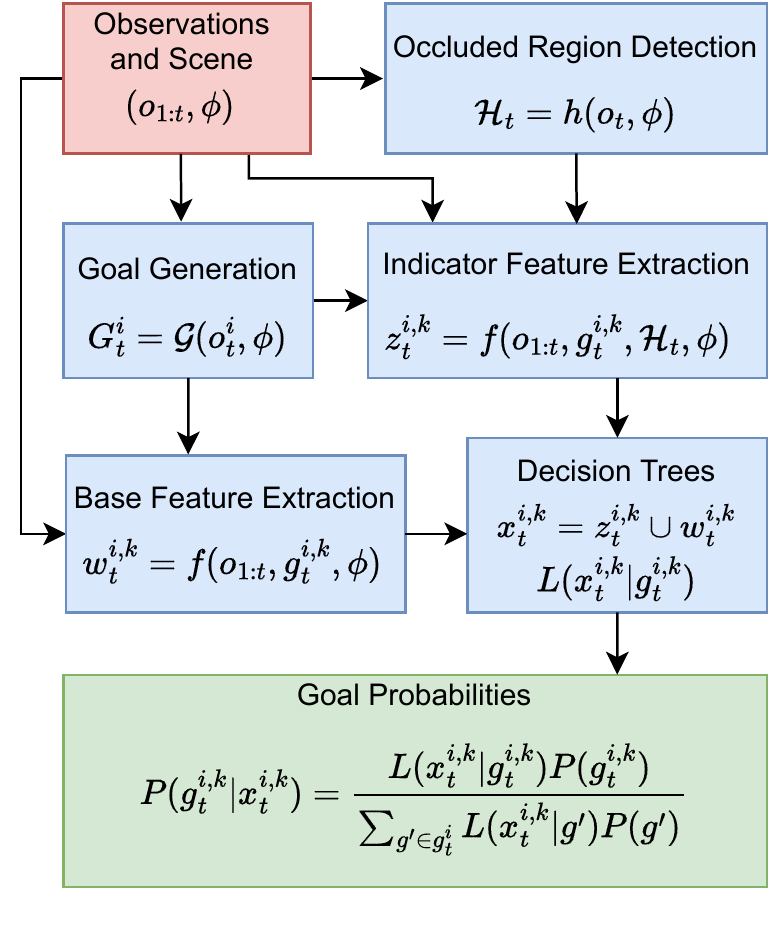} 
  \caption{OGRIT inference system. Red is input and green is output. The arrows signify information flow. Detailed information on each component is given in subsections of \cref{OGRITComponents}.} 
  \label{fig:system-and-tree:system} 
\end{figure}

\section{OGRIT: Goal Recognition with Interpretable Trees under Occlusion} \label{OGRITComponents}

Our method, OGRIT, can infer a probability distribution over possible goals for a vehicle in a partially observable environment with missing data due to occlusions. As shown in \cref{goal_prob_overview}, OGRIT computes the Bayesian posterior probability of each goal $P(g_t^{i,k}|o_{1:t},\phi)$, given a likelihood $L(o_{1:t}|g_t^{i,k},\phi)$ and prior probability $P(g_t^{i,k}|\phi)$ for each goal:
\begin{equation} \label{goal_prob_overview}
P(g_t^{i,k}|o_{1:t},\phi) = \frac{L(o_{1:t}|g_t^{i,k},\phi) P(g_t^{i,k}|\phi)}{\sum_{g^\prime \in g_t^{i,k}}{L(o_{1:t}|g^\prime,\phi) P(g^\prime|\phi)}}
\end{equation}

Similar to the existing GRIT method~\cite{brewitt_grit_2021}, OGRIT computes the likelihoods $L(o_{1:t}|g^{i,k}_t,\phi)$ via decision trees trained from vehicle trajectory data. However, the decision trees used for OGRIT use a novel training algorithm described in  \cref{sec:dt_training} which allows them to handle missing features, while still being highly efficient, interpretable and verifiable.

An overview of the inference process can be seen in \cref{fig:system-and-tree:system}. As input, OGRIT takes the static scene information $\phi$ and the observation history $o_{1:t}$. As output, OGRIT gives the posterior distribution over goals $P(g_t^{i,k}|o_{1:t},\phi)$ for vehicle $i$. First, a set of possible goals is generated based on the current state of the vehicle and the road layout. Next, a set of base features is extracted for each goal (\cref{sec:feature_extraction}), for example, whether $i$ is in the correct lane for $g_t^{i,k}$. A set of occluded regions is detected, which represent the sections of the roads in the local area that are occluded from the ego vehicle's point of view due to objects such as buildings and other vehicles. Using the observations along with the occluded regions, a set of indicator features is extracted. Each indicator feature indicates whether certain base features cannot be inferred due to occlusions. The base features and indicator features are concatenated together to obtain the full set of features. These features are given as input to the associated decision tree for each goal, which outputs the likelihood for that goal. Using the goal likelihoods and prior probabilities, the Bayesian posterior probability for each goal is then computed using \cref{goal_prob_overview}. The following subsections will detail each of these components.

\subsection{Goal Generation}

For each observable vehicle $i$ at time $t$, we generate a set of possible goals $G^i_t$ using a goal generator function $\mathcal{G}(o^i_t,\phi)$. The possible goals are generated by starting from the current lane of vehicle $i$ and performing a limited-depth exhaustive search across the directed graph of lane connections until the nearest junction exits, roundabout exits, or visible lane ends are found, and added to the set of possible goals. We assume that the ego vehicle has access to HD road maps and so can generate goals even in locations that are occluded. Each goal $g_t^{i,k} \in G^i_t$ is assigned a goal type $\tau^{i,k}_t \in \mathcal{T}$, and in this work we use $\mathcal{T} = \{straight \dash on, cross \dash road, exit \dash left, enter \dash left,$ $exit \dash right, enter \dash right, exit \dash roundabout\}$. We use $enter$ to signify entering a higher-priority road and $exit$ to signify leaving a higher-priority road. For each goal type, we train one decision tree. Training decision trees in this manner allow OGRIT to make inferences for previously unseen scenarios, as each decision tree for a certain goal type can be reused when new goals of that type are encountered. This is one of the main differences with GRIT \cite{brewitt_grit_2021}, which trains distinct DTs for each specific goal position in a given scenario.

\subsection{Feature Extraction} \label{sec:feature_extraction}

At each point in time $t$, for each possible goal $g_t^{i,k}$ of each observable vehicle $i$, a set of interpretable features is extracted. We refer to these as the base features, represented by $w^{i,k}_t=f(o_{1:t}, g_t^{i,k}, \phi)$. These features can have binary or scalar values, and contain information extracted from the observation history of vehicle $i$ and other vehicles nearby, and the static scene information. The base features we use include the following, which refer to vehicle~$i$ (further details on feature extraction can be found in the code repository\footref{fn1}): \textit{speed; acceleration; angle-in-lane; angle-to-goal; heading-change-1-second; in-correct-lane; path-to-goal-length; junction-heading-change; roundabout-exit-number; distance-to-vehicle-in-front; speed-of-vehicle-in-front; distance-from-oncoming-vehicle; speed-of-oncoming-vehicle, roundabout-uturn, roundabout-slip-road}. Some of the base features may have missing values due to occlusions, and we represent the set of potentially missing features with $\mathcal{M}$. We use $\mathcal{B}$ to represent the set of features that will never have missing values, and so the set of base features is $\mathcal{M} \cup \mathcal{B}$. For example, if vehicle $i$ is observable, then \textit{position} should never have a missing value ($\textit{position} \in \mathcal{B}$), but the area where there could be a vehicle in front of $i$ may sometimes be occluded. In such cases, \textit{speed-of-vehicle-in-front} has a missing value ($\textit{speed-of-vehicle-in-front} \in \mathcal{M}$) as the ego cannot infer whether there is such vehicle in front of vehicle~$i$.

\subsection{Occlusion Detection} \label{sec:occlusion_detection}
\begin{figure}[t] 
  \begin{subfigure}[b]{0.5\linewidth}
    \centering
    \includegraphics[width=0.95\linewidth]{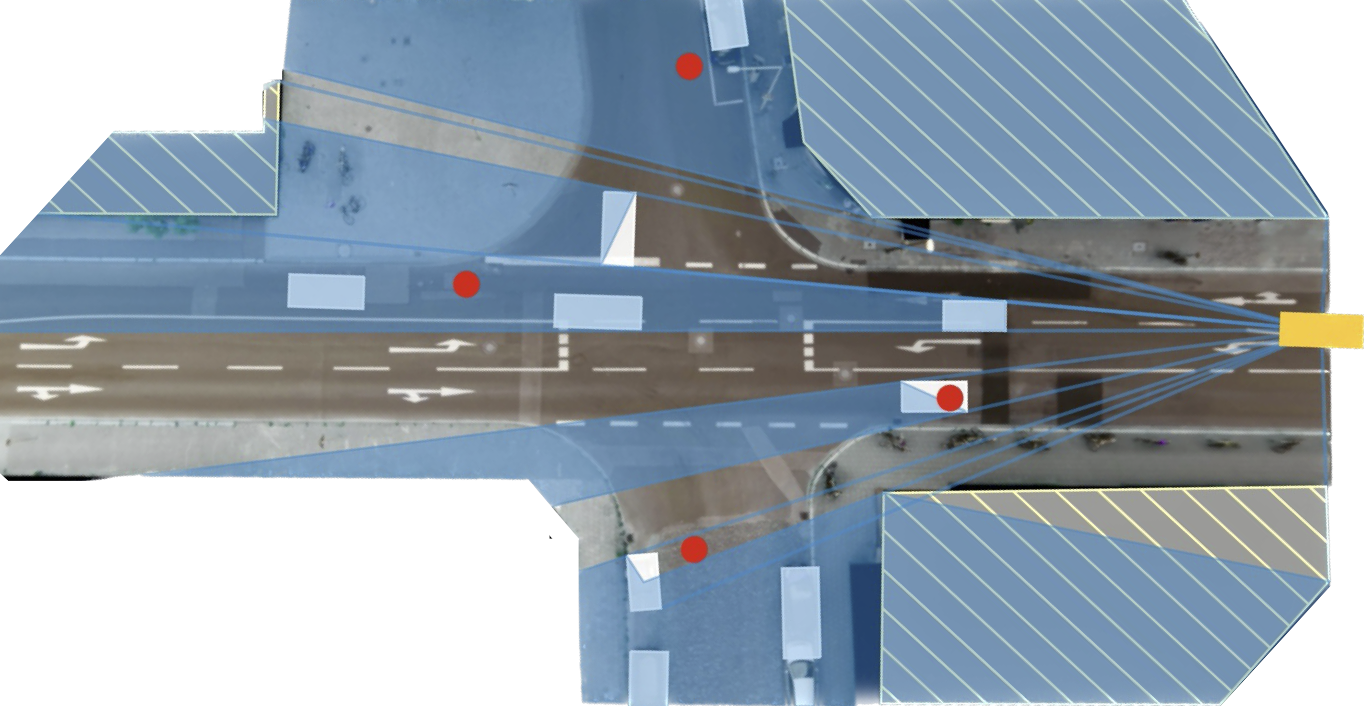} 
    \caption{Bendplatz (inDO)} 
    \label{fig:occlusions:a} 
    \vspace{1ex}
  \end{subfigure}%
  \begin{subfigure}[b]{0.5\linewidth}
    \centering
    \includegraphics[width=0.95\linewidth]{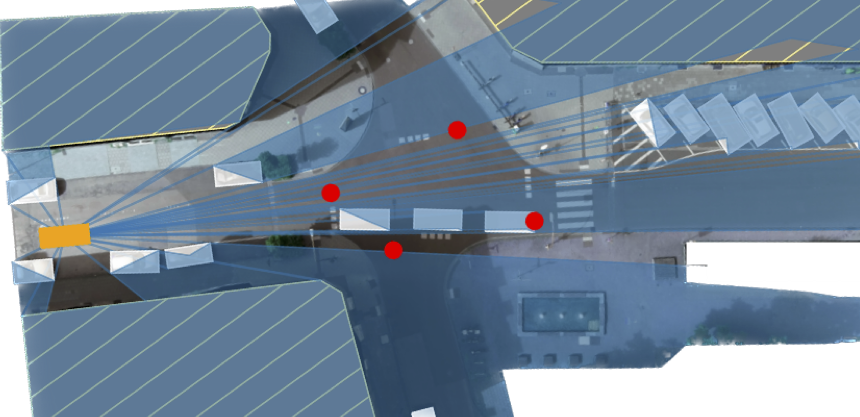} 
    \caption{Frankenburg (inDO)} 
    \label{fig:occlusions:b} 
    \vspace{1ex}
  \end{subfigure} 
  \begin{subfigure}[b]{0.5\linewidth}
    \centering
    \includegraphics[width=0.95\linewidth]{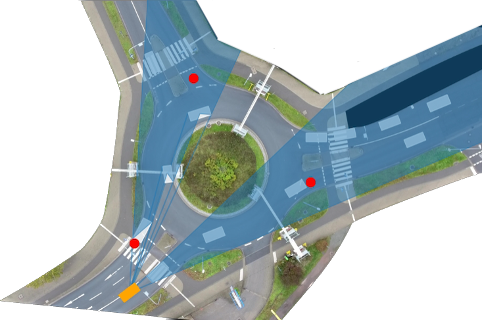} 
    \caption{rdb1 (OpenDDO)} 
    \label{fig:occlusions:c} 
  \end{subfigure}%
  \begin{subfigure}[b]{0.5\linewidth}
    \centering
    \includegraphics[width=0.95\linewidth]{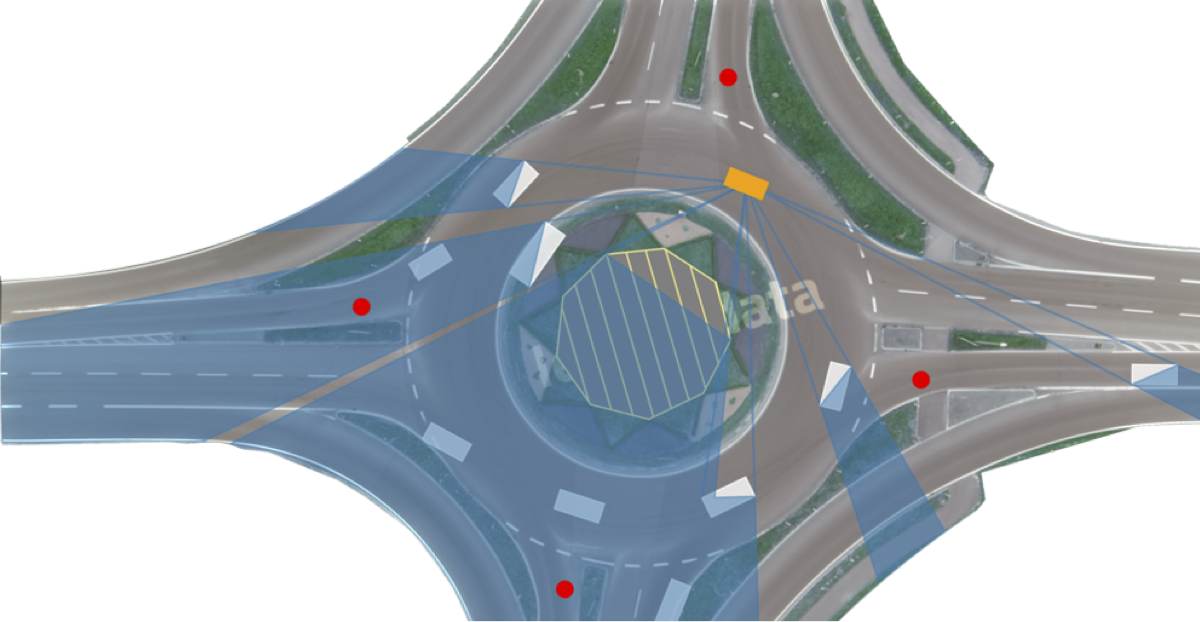} 
    \caption{Neuweiler (rounDO)} 
    \label{fig:occlusions:d} 
  \end{subfigure} 
  \caption{Detected occlusions in snapshots of each scenario. The ego vehicle is shown in orange, traffic agents are shown in white, and buildings are shown in grey with yellow stripes. Goal locations are shown by red dots.}
  \label{fig:occlusions} 
\end{figure}

At each time $t$, we extract the set of regions $\mathcal{H}_t$ which are occluded from the point of view of the ego vehicle using the occlusion detector function $h(o_t, \phi)$. For occlusion detection, we use a two-dimensional top-down representation of the scene, with obstacles such as cars and buildings represented by polygons, as shown in \cref{fig:occlusions}. For each obstacle $u$, we find a pair of line segments $l_1$ and $l_2$ from the centre of the ego vehicle to vertices $v_1$ and $v_2$ of $u$, such that $l_1$ and $l_2$ yield the greatest angle between them. We then extend $l_1$ and $l_2$ so that their total length from the ego vehicle is 100 meters, obtaining new endpoints $v_3$ and $v_4$. We declare the area confined by $v_1$, $v_2$, $v_3$, $v_4$ to be occluded by $u$. Note that the method does not find occluded vehicles, but rather areas in which the ego cannot determine the presence or absence of such vehicles. All areas more than 100 meters from the ego vehicle are considered occluded, as we assume sensors have a limited field of view. We consider a vehicle to be occluded only if its entire boundary is inside an occluded region.

\subsection{Handling Missing Features} \label{sec:missing_features}
The set of potentially missing features $\mathcal{M}$ consists of: \textit{roundabout-exit-number; speed; acceleration; heading-change-1-second; distance-to-vehicle-in-front; speed-of-vehicle-in-front; distance-from-oncoming-vehicle; speed-of-oncoming-vehicle}. We also use a set of indicator features $\mathcal{C}$ which indicate which features are currently missing. The indicator features are binary features which are true if the corresponding base features are missing, or, in other words, if there is an occlusion that prohibits the ego from being able to infer a value for the base feature. For example, the \textit{speed-of-oncoming-vehicle-missing} indicator feature is true if there is an occlusion in the area in which there \textit{could} be an oncoming vehicle, and the ego cannot therefore assess the presence or absence of such vehicle and, thus, of its speed. \textit{Speed-of-oncoming-vehicle-missing} is false when there is no occlusion in that area and the ego can clearly see whether there is an oncoming vehicle and determine its speed. While the base feature \textit{speed-of-oncoming-vehicle} is related to a specific vehicle (for which we require the speed), the indicator feature \textit{speed-of-oncoming-vehicle-missing} is not specific to any vehicle but tells us whether the ego can determine the absence or presence of an oncoming vehicle.

 The function $ind:\mathcal{M} \mapsto \mathcal{C}$ gives the indicator feature for a potentially missing base feature. At each point in time $t$, for each observable vehicle $i$ and possible goal $g_t^{i,k}$, the set of indicator feature values $z^{i,k}_t=f(o_{1:t}, g_t^{i,k},\mathcal{H}_t, \phi)$ is extracted.
The set of all features used for training and inference is given by $\mathcal{L} = \mathcal{B} \cup \mathcal{M} \cup \mathcal{C}$. These features are used in learned DTs as shown in \cref{fig:system-and-tree:tree}.

\subsection{Decision Trees}

\begin{figure}[htp]
    \centering
    \includegraphics[width=0.45\textwidth]{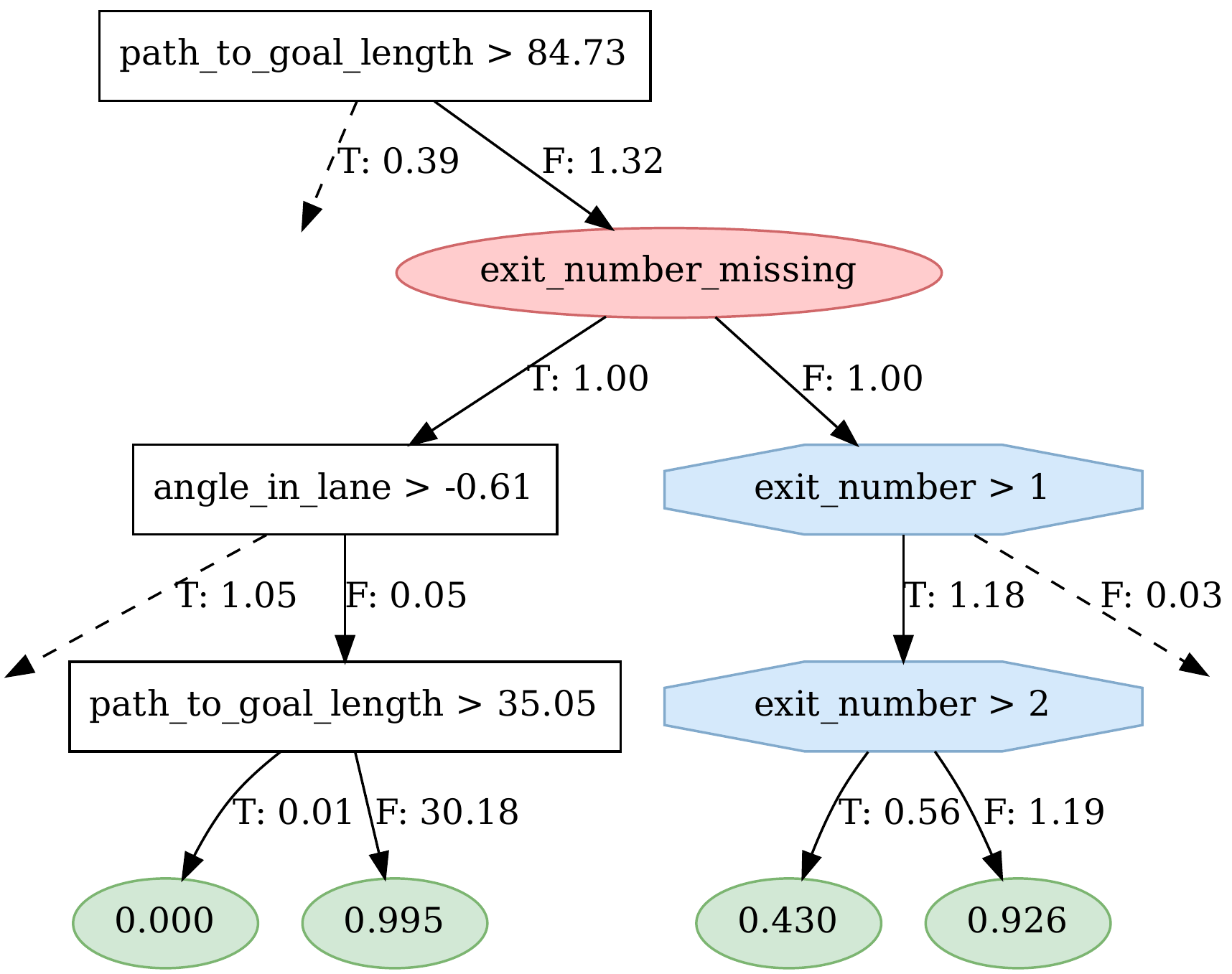} 
    \caption{Illustrative learned DT for \textit{exit-roundabout} goal type: dashed edges point to truncated subtrees. Red oval node represents an indicator feature. Blue octagonal nodes represent potentially missing features (see \cref{sec:missing_features}). The goal likelihood at each leaf node is calculated using the multiplicative weights at each edge and an initial likelihood of 0.5.} 
    \label{fig:system-and-tree:tree} 
\end{figure}

For each goal type $\tau \in \mathcal{T}$, we train a single DT to be used across multiple scenarios. Each decision tree takes the set of features $x^{i,k}_t = z^{i,k}_t \cup w^{i,k}_t$ as input, and outputs a goal likelihood $L(x^{i,k}_t|g_t^{i,k})$. These likelihood values are combined with prior probabilities for each goal to obtain posterior goal probabilities, as shown in \cref{goal_prob_overview}. For simplicity, we use uniform prior probabilities. A weight is assigned to each edge in the decision tree, as shown in \cref{fig:system-and-tree:tree}. The likelihood output at each leaf node of the decision tree is calculated by starting with an initial likelihood of 0.5 at the root node, and then taking the product of this with weights of the edges traversed. Each node $n$ in a decision tree contains a condition which is an inequality on a scalar feature, or the value of a binary feature. To handle missing feature values, potentially missing features $l \in \mathcal{M}$ are only included in decision nodes if the indicator feature value $z_{ind(l)}$ is known to be false.

\subsection{Decision Tree Training} \label{sec:dt_training}

Similar to DT training methods such as CART~\cite{breiman_classification_1984}, ID3~\cite{quinlan_induction_1986} and C4.5~\cite{quinlan_c45_1993}, the DTs used by OGRIT are trained in a top-down manner, starting from the root node and iteratively expanding the tree by adding more decision nodes. However, the OGRIT DT training algorithm introduces some novel elements to handle features with missing values. The three main novelties are: 1) Adding indicator features which indicate that base features are missing; 2) Look-ahead by one level when considering indicator features during training; 3) Adding a term to the cost function when looking ahead to penalise the additional complexity.

\begin{algorithm}[t]
    \fontsize{8.5}{10.0}\selectfont
    
	\textbf{Input:} dataset $D$\\
	\textbf{Returns:} decision tree $T$
	
	\begin{algorithmic}[1]
		\algsetup{linenodelimiter=: }
		\STATE Indicator features with base features missing $\mathcal{C}_t \gets \emptyset$
		\STATE Indicator features with base features known $\mathcal{C}_f \gets \emptyset$
		\STATE $T \gets$ decision tree with root node $n$
		\STATE $recursivelygrowtree(n, D, \mathcal{C}_t, \mathcal{C}_f)$ \COMMENT{See Algorithm 2}
		\STATE $T \gets prune(T, \lambda)$
		\STATE Return $T$
	\end{algorithmic}
	\caption{Train Decision Tree}
	\label{alg:traintree}
\end{algorithm}

\begin{algorithm}[t]
    \fontsize{8.5}{10.0}\selectfont
    
	\textbf{Input:} leaf node $n$ on tree $T$, dataset $D_n$, true indicator features $\mathcal{C}_t$, false indicator features $\mathcal{C}_f$
	
	\begin{algorithmic}[1]
		\algsetup{linenodelimiter=: }
		\IF{$\neg terminate(D_n)$}
		\STATE $\Delta q_{n} \gets 0$ \COMMENT{Impurity (entropy) change for best decision rule}
    		\FORALL{$l | l \in \mathcal{L} \setminus \mathcal{M} \vee ind(l) \in \mathcal{C}_f$}
    		    \STATE $c_n^l \gets \argmax_c ( impuritydecrease(c, l, D_n))$
    		    \STATE $\Delta q \gets -impuritydecrease(c_n^l, l, D_n)$
        		    \IF{$\Delta q < \Delta q_{n}$}
        		        \STATE $c_n, q_{n}, l_{n} \gets c_n^l, q, l$
        		    \ENDIF
    		\ENDFOR
    		\FORALL{$l | l \in \mathcal{M} \wedge ind(l) \notin \mathcal{C}_t \cup \mathcal{C}_f$}
                \STATE $D_{nf} \gets \{(x_j, y_j) | \neg x_{j,ind(l)}  \wedge (x_j, y_j) \in D_n\}$
                \STATE $c_{nf} \gets \argmax_c ( impuritydecrease(c, l, D_{nf}))$
                \STATE $\Delta q \gets - (impuritydecrease(0.5, ind(l), D_n)\;\;\;\;\;\;\;\;\;\;\;\;\;\;\;\;\; $ 
                       $\hphantom \;\;\;\;\;\;\;\;\;\; +  impuritydecrease(c_n^l, l, D_{nf}) - \lambda)$
                \IF{$\Delta q < \Delta q_{n}$}
    		        \STATE $c_n, \Delta q_{n}, l_{n} \gets 0.5, \Delta q, ind(l)$ \COMMENT{0.5 to split binary values}
    		    \ENDIF
    	    \ENDFOR
    		\IF{$\Delta  q_n < 0$}
    		\STATE $D_{nt} \gets \{(x_j, y_j) | x_{j,l_n} > c_n \wedge (x_j, y_j) \in D_n\}$
    		\STATE $D_{nf} \gets D_n \setminus D_{nt}$
    		\IF{$l_n \in \mathcal{C}$}
    		    \STATE $\mathcal{C}_t' \gets \mathcal{C}_t \cup \{l_n\}$ \COMMENT{Indicator features with base features missing for True child}
    		    \STATE $\mathcal{C}_f' \gets \mathcal{C}_f \cup \{l_n\}$ \COMMENT{Indicator features with base features known for False child}
    		\ELSE
    		    \STATE $\mathcal{C}_t' \gets \mathcal{C}_t$
    		    \STATE $\mathcal{C}_f' \gets \mathcal{C}_f$
    		\ENDIF
		    \STATE $recursivelygrowtree(truechild(n), D_{nt}, \mathcal{C}_t', \mathcal{C}_f)$
    		\STATE $recursivelygrowtree(falsechild(n), D_{nf}, \mathcal{C}_t, \mathcal{C}_f')$
    		\ENDIF
		\ENDIF
	\end{algorithmic}
	\caption{Recursively Grow Tree}
	\label{alg:growtree}
\end{algorithm}

Pseudocode for DT training is shown in \cref{alg:traintree}. Decision nodes are added recursively as shown in \cref{alg:growtree} until a termination condition is met, including reaching the maximum depth, the number of samples reaching that node falling below a threshold, or reaching an impurity (entropy) of zero. We use $impuritydecrease(c, l, D_n)$ to refer to the decrease in impurity when adding the decision rule $x_l > c$ to node $n$ of the decision tree, where $D_n$ is the set of training samples reaching node $n$, $c$ is a threshold on $x_l$, and $x_l$ is the value of feature $l$. We prune the tree using cost complexity pruning (CCP)~\cite{breiman_classification_1984}. Decision nodes are iteratively pruned to minimise the cost function shown in \cref{ccp}, where $q_n$ is the impurity at node $n$, $T$ is the set of leaf nodes, and $\lambda > 0$ is a parameter penalising the complexity of the tree:

\begin{equation} \label{ccp}
C(T)=\sum_{n \in T}q_n+\lambda|T|
\end{equation}

To ensure that no nodes are reached where the relevant feature has a missing value, we only allow decision rules relating to potentially missing features to be added to nodes where the relevant indicator feature is found to be false in an ancestor node, meaning that there are no occlusions that prevent the ego from computing the value of those features. When using DT training algorithms such as CART, nodes are expanded one at a time and the decision rule which achieves the maximum impurity decrease is chosen, without taking subsequent child decision nodes into account. However, if we naively tried to add decision rules in this way with indicator features and potentially missing features, it may give unsatisfactory results. There may be cases where adding a potentially missing feature leads to a large impurity decrease, but adding the corresponding indicator feature does not lead to any impurity decrease. As the indicator feature must be added before the potentially missing feature can be added, in such a case the potentially missing feature would never be added, despite its effectiveness. To overcome this problem, we look ahead by one level when considering indicator features and missing features, rather than using the greedy approach of only considering one decision node at a time. We consider a parent which uses the indicator feature and a child which uses the potentially missing feature. As this operation considers adding two nodes to the decision tree rather than the usual one node, there should be an additional complexity penalty added to the cost function relative to when we consider adding only one decision node. The same $\lambda$ complexity penalty used during CCP can be used as a penalty for adding two decision nodes at once, as shown in line 11 of \cref{alg:growtree}.

We train one DT for each goal type, and each DT is trained using a set of samples whose possible goal $g_t^{i,k}$ has the given goal type. These make up the dataset $D=\{(x_1, y_1), ..., (x_N, y_N)\}$, where $(x_j, y_j)$ is a sample, $x_j$ is the set of features input to the DT, and $y_j$ is a Boolean value that indicates whether the possible goal is the true goal for the vehicle. $x_{j,l}$ represents the value of feature $l \in \mathcal{L}$ for sample $j$. We use $D_n \subseteq D$ to represent the set of samples which reach node $n$ of the DT. The likelihood value assigned to each node is calculated using several sample counts, each of which is regularised using Laplace smoothing to obtain pseudo-counts. These include the number of samples for which the possible goal is the true goal, $N_{G}=|\{j | y_j \}|$, the number of samples which reach node $n$ and the possible goal is the true goal, $N_{nG}=|\{j | (x_j, y_j) \in D_n \wedge\ y_j\}|$, and the number of samples at node $n$ where the possible goal is not the true goal, $N_{n \bar{G}}=|D_n| - N_{nG}$. In some cases, there may be a large imbalance between the number of samples where the possible goal is and is not the true goal. To compensate for this, we weight the samples where the possible goal is the true goal by $w_G=N/N_G$, and weight the samples where the possible goal is not the true goal by $w_{\bar{G}}=N/N_{\bar{G}}$, where N is the total number of samples. The likelihood value at each node is then calculated as shown in \cref{likelihood}:

\begin{equation} \label{likelihood}
    L_n=\frac{w_G N_{nG}}{w_G N_{nG}+w_{\bar{G}} N_{n\bar{G}}}
\end{equation}

After the likelihood values $L_n$ at each node are calculated, the weights for each edge between a child and parent node are calculated as $L_{child} / L_{parent}$.

\section{Evaluation}
\label{sec:evaluation}

We compare the performance of OGRIT to several baselines across three vehicle trajectory datasets. We show that OGRIT can achieve fast, accurate and interpretable inference under occlusion. We run formal verification for several propositions about inferences made by OGRIT.

\subsection{Datasets}

To train and evaluate our models, we use three datasets: the \emph{inDO}, \emph{rounDO} and \emph{OpenDDO} datasets\footref{fn1}. We generate these datasets by using the occlusion detector described in \cref{sec:occlusion_detection} to extract occluded regions in the inD~\cite{bock_ind_2019}, rounD~\cite{krajewski_round_2020}, and OpenDD~\cite{openDD_dataset}. The inD, rounD and OpenDD datasets consist of vehicle trajectories extracted from videos taken by drones hovering above junctions or roundabouts. We use three scenarios from the inD dataset, named Heckstrasse, Bendplatz and Frankenburg, one scenario from the rounD dataset, named Neuweiler, and all scenarios from the OpenDD dataset, rdb1 to rdb7. The data for each scenario is divided into continuous recordings or episodes. In each scenario in the inDO dataset, we hold out one randomly chosen recording for testing, one recording for validation, and use the remaining recordings for training. Due to the larger number of recordings in the other two datasets, we hold out three recordings for validation and testing in roundDO and perform a 60:20:20 training-validation-test split for OpenDDO. During each recording, we extract samples at one second intervals. At each interval, we consider each vehicle as an ego vehicle and extract samples for each of the other target vehicles that are observable by the ego vehicle up to the point when each target vehicle reaches its goal.

\subsection{OGRIT Implementation}  \label{sec:dt-training-details}

We use the following hyperparameter values for decision tree training. The cost complexity pruning parameter $\lambda = 0.0001$ was selected by a grid search to maximise the true goal probability of OGRIT on the validation set. The maximum decision tree depth is set to 7, to ensure that the learned trees are interpretable. To reduce overfitting, the minimum number of training samples allowed at each leaf node is set to 10. For Laplace smoothing of sample counts, a value of $\alpha=1$ is used. We use uniform prior probabilities for the goal distributions.

\begin{figure}[t]
    \centering
    \includegraphics[width=0.48\textwidth]{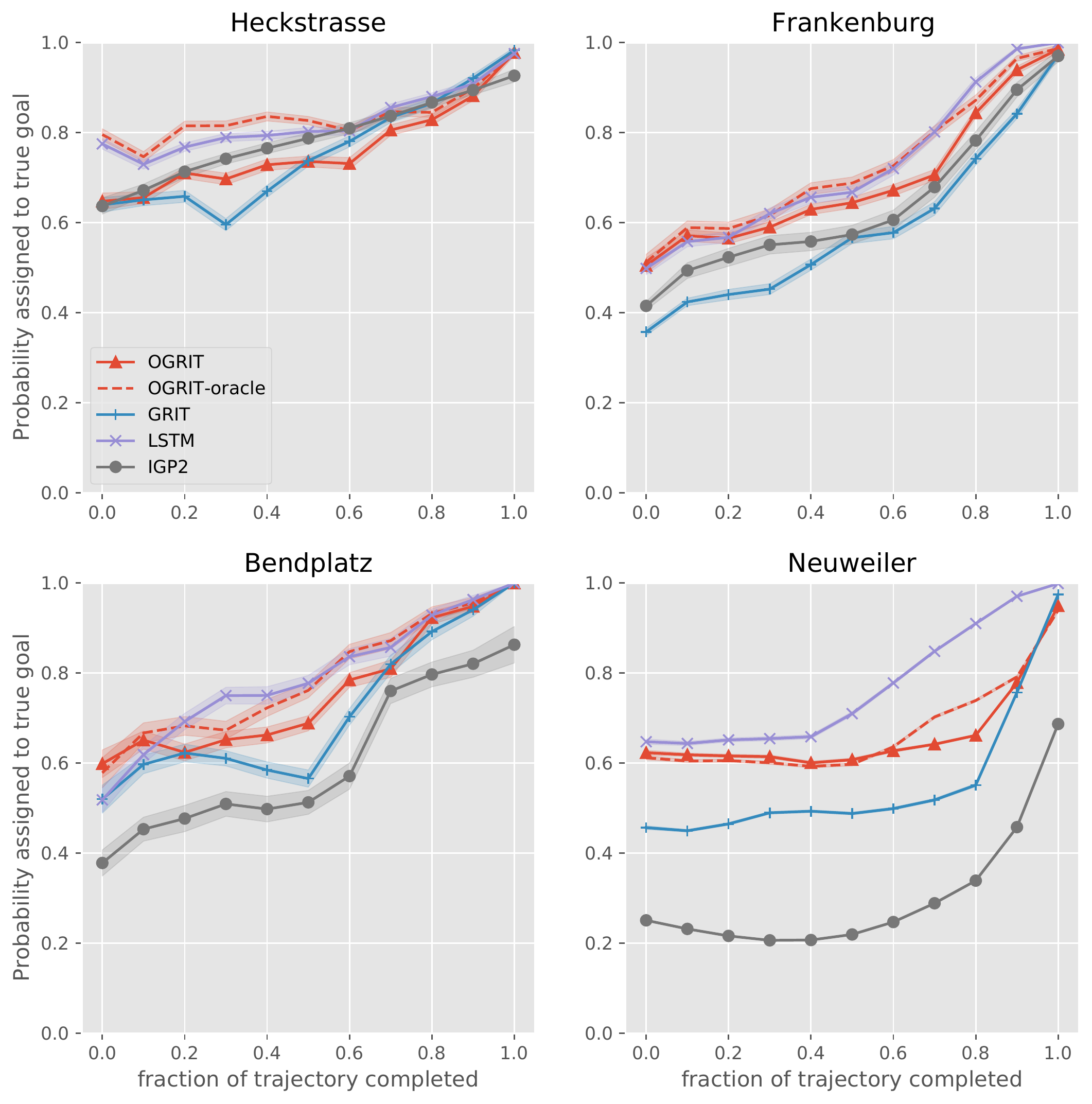}
    \caption{Average probability assigned to the true goal by OGRIT, OGRIT-Oracle which has access to all occluded information, GRIT with potentially missing features left out, an LSTM neural network, and inverse planning method IGP2. Fraction of trajectory completed is the fraction of time passed between the first observation of a vehicle and reaching its goal. Shaded areas show standard error.}
    \label{fig:ogrit_goalprob}
\end{figure}

\subsection{Baselines}

\textbf{OGRIT-Oracle:} A version of OGRIT that has access to all occluded information. This method acts as an upper bound to the performance of OGRIT. For this baseline, the same hyperparameters were used as for OGRIT, described in \cref{sec:dt-training-details}. During training and evaluation, the values of all features missing due to occlusions were provided to the model. The training process was also modified to eliminate the influence of potentially missing features. As a result, potentially missing features could be added to trees even if the corresponding indicator feature had not been added in an ancestor node. 

\textbf{GRIT:} The existing GRIT \cite{brewitt_grit_2021} method, which only uses features from the set $\mathcal{B}$ which are never missing due to occlusions. This method has specialised DTs trained for each scenario. Similarly to OGRIT, we restrict the maximum tree depth to 7 for fair comparison. A different cost complexity pruning parameter $\lambda$ was selected by grid search for each scenario. We do not allow potentially missing features to be added to the DTs during training, as GRIT has no way of handling features with missing values.

\textbf{LSTM:} Long Short-Term Memory (LSTM) neural network \cite{lstm}.  As input, LSTM takes the sequence of features used by OGRIT for the target vehicle. Whenever one of the features is missing, we fill the input value corresponding to the feature with a -1. We trained one LSTM for each goal type, which predicts the likelihood that the sequence of features belongs to a specific goal location with the goal type of that LSTM. We use a single LSTM layer, where each hidden unit has cell size of 64. The outputs of the LSTM layer are passed through a fully connected layer with 725 hidden units. 

\textbf{IGP2:} In addition to learning-based GR methods, we also use the inverse planning-based GR method from IGP2~\cite{albrecht_interpretable_2021}. This method uses a planner to find the optimal plan for a goal from both the initially observed and current position of the target vehicle. The cost difference between the two plans is then used to calculate the goal likelihood. IGP2 can handle occlusions in observed vehicle trajectories by using a planner to fill sections of the trajectory missing due to occlusions, as detailed in ~\cite{albrecht_interpretable_2021}. We use the same maneuvers and macro actions as in the original work, except for \textit{Stop}.

We used the following parameter values across all datasets: give way distance of 15; give way lane angle threshold of $\pi / 6$; give way turn target threshold of 1; maneuver point spacing of 0.25; maneuver max speed of 10; maneuver min speed of 3; switch lane minimum switch length of 10. For the full set of parameters, please refer to the code repository.\footref{fn1}

For the velocity smoother used in IGP2, we used parameter values of $v_{max}= speed limit$, $a_{max}=5$, and $\Delta_t=0.1$. Similarly to \cite{brewitt_grit_2021}, we use sum of squares instead of L2 norm to improve velocity smoother convergence. The parameters used in velocity smoother are the same in all datasets.

\begin{figure}[t]
    \centering
    \includegraphics[width=0.32\textwidth]{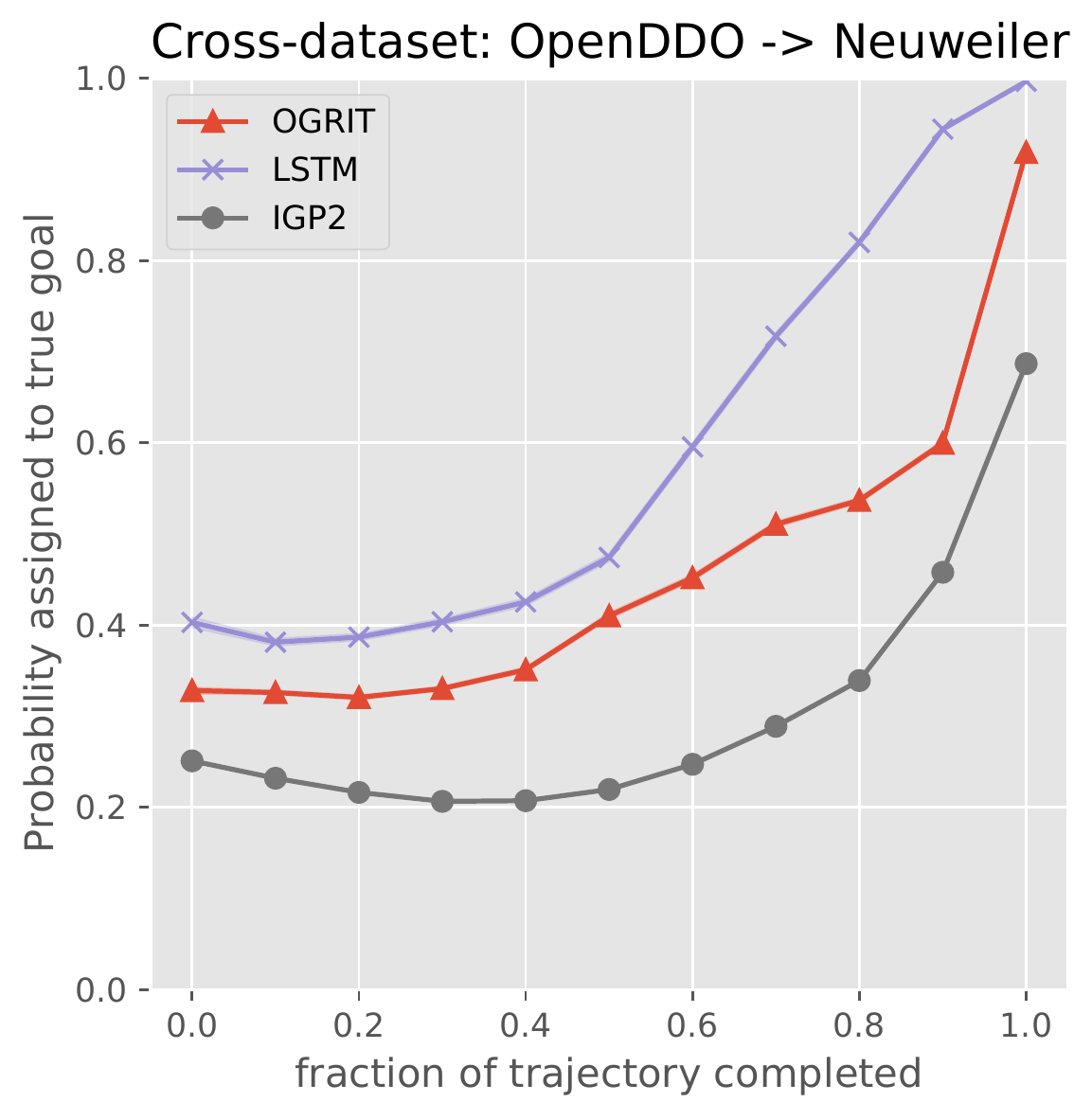}
    \caption{Average probability assigned to the true goal by OGRIT, LSTM, and IGP2, when generalising between different datasets. In this case, the models were trained on the OpenDDO dataset, and evaluated on the Neuweiler scenario from the rounDO dataset. Fraction of trajectory completed is the fraction of time passed between the first observation of a vehicle and reaching its goal.}
    \label{fig:ogrit_cross_dataset_generalisation}
\end{figure}

\subsection{Goal Recognition Accuracy Within Datasets} \label{sec:generalisation-within-scenarios}

In this section, we discuss the results from our first experiment, where we train on a subset of episodes from each scenario in the inDO and rounDO datasets, and then evaluate on a set of held out episodes from the same scenarios. The average probability assigned to the ground truth goal by OGRIT and the baselines is shown in \cref{fig:ogrit_goalprob}. In all scenarios, OGRIT achieves higher accuracy than GRIT. OGRIT also uses the same learned DTs to generalise across all scenarios, while GRIT requires specialised DTs for each scenario. The accuracy of OGRIT is close to that of the OGRIT-Oracle, despite not having access to occluded information. In all scenarios other than Heckstrasse, OGRIT achieves better performance than IGP2. This could be due to IGP2's strict assumption that vehicles use maneuvers from a predefined library. In addition, OGRIT can use information from the initial observed vehicle state to make inferences about goal likelihood, while IGP2 only takes into account the vehicle's actions after its initial state. LSTM achieves the highest overall accuracy. One reason for this may be that LSTM receives the entire sequence of historical feature values as input, while OGRIT only receives the features from the current timestep. However, the LSTM is not interpretable or verifiable due to its large number of learned parameters.

The largest performance difference between OGRIT and GRIT is found in the Neuweiler scenario, which contains a roundabout with four exits. In this scenario, we found that the potentially missing feature \textit{exit-number} has a large impact on the inferences made. The \textit{exit-number} for a certain \textit{exit-roundabout} goal $g_t^{i,k}$ represents the number of roundabout exits that would be passed by the vehicle $i$ between entering the roundabout and reaching $g_t^{i,k}$. Knowledge of the \textit{exit-number} is important for goal recognition, as vehicles rarely exit at the same place that they entered and rarely take the first exit due to the alternative slip roads seen in \cref{fig:occlusions:d}.

\subsection{Cross-Dataset Generalisation}

We investigate the ability of OGRIT to generalise to a new dataset, unseen during training. In this experiment, we trained OGRIT and the baselines on all scenarios from the openDDO dataset, and then evaluated on the Neuweiler scenario from the rounDO dataset. For these experiments, we used only one feature, \textit{angle-to-goal} when training OGRIT, which was selected due to its generalisation effectiveness. \cref{fig:ogrit_cross_dataset_generalisation} shows the average probability assigned to the ground truth goal by each of the models.

Overall, LSTM achieves a higher accuracy than OGRIT, especially when more than 0.6 of the vehicle's trajectory has been completed. As was noted earlier, one reason for this may be that LSTM is given the entire sequence of historical feature values as input, while OGRIT only receives the feature values from the current timestep. In addition, LSTM has a much larger number of parameters, which may achieve a better fit for the data. OGRIT still achieves reasonable accuracy, and significantly outperforms IGP2. IGP2 achieves the same results as described in \cref{sec:generalisation-within-scenarios}, as it is not trained on data, so the change in training data has no effect on results.

\subsection{Inference Speed}
We measure the average time taken for OGRIT to compute the posterior probability over goals, based on our Python implementation and measured on an AMD EPYC 7502 CPU. This includes the entire inference process shown in \cref{fig:system-and-tree:system}. The average inference time of OGRIT was 26 ms. This is fast enough to make inferences in real time to be used as input to a prediction and planning module. For comparison, GRIT had a faster inference time of 4.9 ms, due to the lack of occlusion detection and indicator feature extraction.

\subsection{Interpretability}

We found that DTs learned by OGRIT are easily interpretable. For example, consider the DT for \textit{exit-roundabout} goals shown in \cref{fig:system-and-tree:tree}. If the leftmost leaf node shown in \cref{fig:system-and-tree:tree} is reached during inference, we can infer that: ``The \textit{exit-roundabout} goal has a likelihood of 0.0003, because the path to goal length is less than 84.73 metres (weight 1.32), the angle in lane is greater than 0.61 radians to the right (weight 0.05), and the path to goal length is greater than 35.05 metres (weight  0.01)''. The low likelihood makes sense in this case because the vehicle has turned harshly to the right, while still over 35 metres from the relevant roundabout exit.

\subsection{Verification}

To perform formal verification, the trained model and a proposition to be verified $\Psi$ are first represented using propositional logic, as described in \cite{brewitt_grit_2021}. Next, we can verify that $\Psi$ will always hold true by using a satisfiability modulo theories (SMT) solver to prove that $\neg \Psi$ is unsatisfiable. In our case, we use the Z3 solver~\cite{de_moura_z3_2008}. In the case that verification fails, the SMT solver provides a counterexample which can teach us more about how our model works. We ran verification for three propositions on the trained OGRIT model. We use $x^g_t$ to represent the set of all feature values for goal $g$ at timestep or scene instance $t$. The value of feature $l \in \mathcal{L}$ is represented by $x^{g,l}_t$. We use feature identifiers: $onm$ (\textit{oncoming-vehicle-missing}), $xn$ (\textit{exit-number}), $xnm$ (\textit{exit-number-missing}), $pth$ (\textit{path-to-goal-length}), $an$ (\textit{angle-in-lane}).

\textbf{Goal distribution entropy with the vehicle in front occluded:}
If certain feature values are missing, a reasonable expectation is that the model should be more uncertain about the goals of vehicles than if the feature values are not missing. In other words, the entropy of the goal distribution should be higher if the feature values are missing. In the case that there are two goals, a decrease in the probability of the most probable goal is equivalent to an increase in entropy. For the case that a vehicle is at a junction where there are \textit{straight-on} and \textit{exit-left} goals possible, we successfully verified the following proposition: ``If the \textit{oncoming-vehicle-missing} indicator feature is true, then the entropy of the goal distribution should be greater than or equal to the case where the \textit{oncoming-vehicle-missing} indicator feature is false, all other features being equal'':
$$\bigwedge_{g \in \{G1,G2\}} ( (x^{g,onm}_{t1} \wedge \neg x^{g,onm}_{t2} )
\bigwedge_{l \in \mathcal{L}, \atop l \neq onm} (x^{g,l}_{t1}=x^{g,l}_{t2})) \implies $$
$$((P(G1|x^{G1}_{t1}) < P(G2|x^{G2}_{t1}) \implies $$ $$P(G2|x^{G2}_{t2}) \geq P(G2|x^{G2}_{t1})) \wedge (P(G1|x^{G1}_{t1}) > P(G2|x^{G2}_{t1}) $$
$$\implies P(G1|x^{G1}_{t2}) \geq P(G1|x^{G1}_{t1})))$$

\textbf{Goal probability with oncoming vehicles occluded:}
If a target vehicle is stopped at a junction entrance, one explanation for stopping may be that its goal is \textit{enter-right}, and there is an oncoming vehicle which would block its way while turning. Even if oncoming vehicles are occluded, it would still be reasonable to assign a high probability to the \textit{enter-right} goal, as there could be an oncoming vehicle hidden from view \cite{hanna_interpretable_2021}. However, if there are no occlusions and the ego vehicle can see that there are no oncoming vehicles, then stopping would be irrational for the \textit{enter-right} goal and \textit{enter-right} should be given a low probability. We run verification of the proposition: ``If a vehicle is stopped at a junction, angled straight ahead in its lane, and oncoming vehicles are occluded, then \textit{enter-right} should have a higher probability than if oncoming vehicles are not occluded and there is no oncoming vehicle'', represented as follows, where $G1$ is the \textit{enter-right} goal:
$$x^{G1,onm}_{t1} \wedge \neg x^{G1,onm}_{t2} \implies P(G1|x^{G1}_{t2}) \geq P(G1|x^{G1}_{t1})$$

In this case, verification failed on the learned model, and the solver provides a counterexample where there is a stopped vehicle 8.5 metres in front of the target vehicle. In such a case, it would be rational for a vehicle with an \textit{enter-right} goal to stop, even if there are no oncoming vehicles.

\textbf{Exit roundabout goal likelihood with roundabout exit number occluded:}
When a vehicle is in a roundabout, it is a reasonable expectation that the goal of exiting to the same road from which a vehicle entered should have a lower likelihood than other exits. In the Neuweiler scenario, this corresponds to taking the fourth exit relative to the entry point. If the exit number for goal $g$ is missing due to occlusions, then the \textit{exit-number} for $g$ could have any value, and it would be reasonable to expect that the goal likelihood should be higher than if the \textit{exit-number} is known to be four. We successfully verified following proposition: ``If the exit number for goal $g$ is known to be four, then the likelihood of $g$ should be lower than or equal to the likelihood of $g$ if the \textit{exit-number} for $g$ is missing due to occlusions, if \textit{path-to-goal-length} is 50m, and \textit{angle-in-lane} is zero, all other features being equal'':
$$x^{g,pth}_{t1} = 50  \wedge x^{g,an}_{t1} = 0 \wedge x^{g,xn}_{t2} = 4 \wedge x^{g,xnm}_{t1} \wedge \neg x^{g,xnm}_{t2} $$
$$\bigwedge_{l \in \mathcal{L}, \atop l \neq xnm} x^{g,l}_{t1}=x^{g,l}_{t2} \implies L(x^{g}_{t1}|g) \geq L(x^{g}_{t2}|g)$$

\section{Conclusion}
\label{sec:conclusion}

We presented a novel autonomous vehicle goal recognition method named OGRIT. We showed that OGRIT can handle missing data due to occlusions and make inferences across multiple scenarios using the same learned model, while resulting in fast, accurate, interpretable and verifiable inference. We also showed that the models learned by OGRIT can generalise to a new scenario in a dataset unseen during training. 

Future directions could include considering occlusions from the point of view of the target vehicle in addition to the ego vehicle. For example, the target vehicle may drive cautiously if an oncoming lane is occluded to the target vehicle, even if the ego vehicle can see that there are no oncoming vehicles, which is not currently handled by OGRIT. OGRIT also assumes a predefined set of goal types, which may be incomplete. This could be addressed in future work through anomaly detection, by declaring the goal "unknown" if the likelihood for all generated goals is low.

\addtolength{\textheight}{-3cm}   %

\bibliographystyle{IEEEtran}
\bibliography{library}

\begin{thebibliography}{10}
\providecommand{\url}[1]{#1}
\csname url@samestyle\endcsname
\providecommand{\newblock}{\relax}
\providecommand{\bibinfo}[2]{#2}
\providecommand{\BIBentrySTDinterwordspacing}{\spaceskip=0pt\relax}
\providecommand{\BIBentryALTinterwordstretchfactor}{4}
\providecommand{\BIBentryALTinterwordspacing}{\spaceskip=\fontdimen2\font plus
\BIBentryALTinterwordstretchfactor\fontdimen3\font minus
  \fontdimen4\font\relax}
\providecommand{\BIBforeignlanguage}[2]{{%
\expandafter\ifx\csname l@#1\endcsname\relax
\typeout{** WARNING: IEEEtran.bst: No hyphenation pattern has been}%
\typeout{** loaded for the language `#1'. Using the pattern for}%
\typeout{** the default language instead.}%
\else
\language=\csname l@#1\endcsname
\fi
#2}}
\providecommand{\BIBdecl}{\relax}
\BIBdecl

\bibitem{rhinehart_precog_2019}
N.~Rhinehart, R.~McAllister, K.~Kitani, and S.~Levine, ``{PRECOG}: {Prediction}
  conditioned on goals in visual multi-agent settings,'' in \emph{{IEEE} ICCV},
  2019.

\bibitem{mangalam_it_2020}
K.~Mangalam, H.~Girase, S.~Agarwal, K.-H. Lee, E.~Adeli, J.~Malik, and
  A.~Gaidon, ``\BIBforeignlanguage{en}{It {Is} {Not} the {Journey} {But} the
  {Destination}: {Endpoint} {Conditioned} {Trajectory} {Prediction}},'' in
  \emph{\BIBforeignlanguage{en}{ECCV}}, 2020.

\bibitem{zhao_tnt_2020}
H.~Zhao, J.~Gao, T.~Lan, C.~Sun, B.~Sapp, B.~Varadarajan, Y.~Shen, Y.~Shen,
  Y.~Chai, C.~Schmid, C.~Li, and D.~Anguelov, ``Tnt: Target-driven trajectory
  prediction,'' in \emph{CoRL}, 2020.

\bibitem{albrecht_interpretable_2021}
S.~V. Albrecht, C.~Brewitt, J.~Wilhelm, B.~Gyevnar, F.~Eiras, M.~Dobre, and
  S.~Ramamoorthy, ``Interpretable goal-based prediction and planning for
  autonomous driving,'' in \emph{IEEE ICRA}, 2021.

\bibitem{butler_infeasibility_1993}
R.~Butler and G.~Finelli, ``The infeasibility of quantifying the reliability of
  life-critical real-time software,'' \emph{TSE}, 1993.

\bibitem{Luckcuck_formal_2019}
M.~Luckcuck, M.~Farrell, L.~A. Dennis, C.~Dixon, and M.~Fisher, ``Formal
  {Specification} and {Verification} of {Autonomous} {Robotic} {Systems},''
  \emph{ACM CSUR}, 2019.

\bibitem{schmidt_can_2021}
L.~M. Schmidt, G.~Kontes, A.~Plinge, and C.~Mutschler, ``Can {You} {Trust}
  {Your} {Autonomous} {Car}? {Interpretable} and {Verifiably} {Safe}
  {Reinforcement} {Learning},'' in \emph{{IEEE} IV}, 2021.

\bibitem{bock_ind_2019}
J.~Bock, R.~Krajewski, T.~Moers, S.~Runde, L.~Vater, and L.~Eckstein, ``The ind
  dataset: A drone dataset of naturalistic road user trajectories at german
  intersections,'' in \emph{{IEEE} IV}, 2020.

\bibitem{krajewski_round_2020}
R.~Krajewski, T.~Moers, J.~Bock, L.~Vater, and L.~Eckstein, ``The round
  dataset: A drone dataset of road user trajectories at roundabouts in
  germany,'' in \emph{IEEE ITSC}, 2020.

\bibitem{openDD_dataset}
A.~Breuer, J.-A. Termöhlen, S.~Homoceanu, and T.~Fingscheidt, ``Opendd: A
  large-scale roundabout drone dataset,'' September 2020.

\bibitem{lee_desire_2017}
N.~Lee, W.~Choi, P.~Vernaza, C.~B. Choy, P.~H.~S. Torr, and M.~Chandraker,
  ``{DESIRE}: {Distant} future prediction in dynamic scenes with interacting
  agents,'' in \emph{{IEEE} {CVPR}}, 2017.

\bibitem{chai_multipath_2019}
Y.~Chai, B.~Sappm, M.~Bansal, and D.~Anguelov, ``{MultiPath}: {Multiple}
  {Probabilistic} {Anchor} {Trajectory} {Hypotheses} for {Behavior}
  {Prediction},'' in \emph{CoRL}, 2019.

\bibitem{Knittel2023dipa}
A.~Knittel, M.~Hawasly, S.~V. Albrecht, J.~Redford, and S.~Ramamoorthy,
  ``{DiPA:} probabilistic multi-modal interactive prediction for autonomous
  driving,'' \emph{IEEE RA-L}, vol.~8, no.~8, pp. 4887--4894, 2023.

\bibitem{casas_intentnet_2018}
S.~Casas, W.~Luo, and R.~Urtasun, ``{IntentNet}: {Learning} to {Predict}
  {Intention} from {Raw} {Sensor} {Data},'' in \emph{CoRL}, 2018.

\bibitem{christianos2023planning}
F.~Christianos, P.~Karkus, B.~Ivanovic, S.~V. Albrecht, and M.~Pavone,
  ``Planning with occluded traffic agents using bi-level variational occlusion
  models,'' in \emph{ICRA}, 2023.

\bibitem{antonello2022flash}
M.~Antonello, M.~Dobre, S.~V. Albrecht, J.~Redford, and S.~Ramamoorthy,
  ``Flash: Fast and light motion prediction for autonomous driving with
  bayesian inverse planning and learned motion profiles,'' in \emph{IEEE/RSJ
  IROS}, 2022.

\bibitem{ayers_parot_2020}
E.~W. Ayers, F.~Eiras, M.~Hawasly, and I.~Whiteside, ``{PaRoT}: {A} {Practical}
  {Framework} for {Robust} {Deep} {Neural} {Network} {Training},'' in
  \emph{NFM}, 2020.

\bibitem{hardy_contingency_2013}
J.~Hardy and M.~Campbell, ``Contingency {Planning} {Over} {Probabilistic}
  {Obstacle} {Predictions} for {Autonomous} {Road} {Vehicles},'' \emph{T-RO},
  2013.

\bibitem{bandyopadhyay_intention-aware_2013}
T.~Bandyopadhyay, K.~S. Won, E.~Frazzoli, D.~Hsu, W.~S. Lee, and D.~Rus,
  ``Intention-aware motion planning,'' in \emph{WAFR {X}}, 2013.

\bibitem{ziebart_maximum_2008}
B.~D. Ziebart, A.~Maas, J.~A. Bagnell, and A.~K. Dey, ``Maximum entropy inverse
  reinforcement learning,'' \emph{AAAI}, 2008.

\bibitem{darweesh_estimating_2019}
H.~Darweesh, E.~Takeuchi, and K.~Takeda, ``Estimating the {Probabilities} of
  {Surrounding} {Vehicles}’ {Intentions} and {Trajectories} using a
  {Behavior} {Planner},'' \emph{IJAE}, 2019.

\bibitem{hanna_interpretable_2021}
J.~Hanna, A.~Rahman, E.~Fosong, F.~Eiras, M.~Dobre, J.~Redford, S.~Ramamoorthy,
  and S.~Albrecht, ``Interpretable goal recognition in the presence of occluded
  factors for autonomous vehicles,'' in \emph{IEEE/RSJ IROS}, 2021.

\bibitem{brewitt_grit_2021}
C.~Brewitt, B.~Gyevnar, S.~Garcin, and S.~V. Albrecht, ``{GRIT:} fast,
  interpretable, and verifiable goal recognition with learned decision trees
  for autonomous driving,'' in \emph{IEEE/RSJ IROS}, 2021.

\bibitem{gavankar_decision_2015}
S.~Gavankar and S.~Sawarkar, ``Decision {Tree}: {Review} of {Techniques} for
  {Missing} {Values} at {Training}, {Testing} and {Compatibility},'' in
  \emph{AIMS}, 2015.

\bibitem{khosravi_handling_2020}
P.~Khosravi, A.~Vergari, Y.~Choi, Y.~Liang, and G.~V. den Broeck, ``Handling
  missing data in decision trees: A probabilistic approach,'' in \emph{ICML
  ARTEMISS}, 2020.

\bibitem{quinlan_induction_1986}
J.~R. Quinlan, ``Induction of decision trees,'' \emph{Machine Learning}, 1986.

\bibitem{friedman_lazy_1997}
J.~Friedman, R.~Kohavi, and Y.~Yun, ``Lazy {Decision} {Trees},'' \emph{AAAI},
  1997.

\bibitem{breiman_classification_1984}
L.~Breiman, \emph{Classification and regression trees}, 1984.

\bibitem{quinlan_c45_1993}
J.~R. Quinlan, \emph{C4.5: programs for machine learning}.\hskip 1em plus 0.5em
  minus 0.4em\relax Morgan Kaufmann Publishers Inc., 1993.

\bibitem{lstm}
S.~Hochreiter and J.~Schmidhuber, ``Long short-term memory,'' \emph{Neural
  computation}, 1997.

\bibitem{de_moura_z3_2008}
L.~de~Moura and N.~Bjørner, ``\BIBforeignlanguage{en}{Z3: {An} {Efficient}
  {SMT} {Solver}},'' in \emph{\BIBforeignlanguage{en}{TACAS}}, 2008.

\end{thebibliography}

\end{document}